\documentclass{article}

\usepackage{iclr2021_conference,times}

\usepackage[utf8]{inputenc} 
\usepackage[T1]{fontenc}    
\usepackage{hyperref}       
\usepackage{url}            
\usepackage{booktabs}       
\usepackage{amsfonts}       
\usepackage{nicefrac}       
\usepackage{microtype}      
\usepackage{xcolor}         
\usepackage{authblk}
\usepackage{tabularx}
\usepackage{xspace}
\usepackage{amsmath}
\usepackage{graphicx}
\usepackage{algpseudocode}
\usepackage{algorithm}
\usepackage{multirow}

\usepackage{multirow}

\title{\centering Model Agnostic Hybrid Sharding for Heterogeneous Distributed Inference}

\author{%
    Claudio Angione, Yue Zhao, Harry Yang, Ahmad Farhan,  
    Fielding Johnston, \\[0.2em]
    James Buban,
    Patrick Colangelo
}

\affil[]{Nesa Research}
\affil[]{\texttt{research@nesa.ai}}
\begin{document}

\maketitle

\begin{abstract}
The rapid growth of large-scale AI models, particularly large language models has brought significant challenges in data privacy, computational resources, and accessibility. Traditional centralized architectures often struggle to meet required data security and scalability needs which hinders the democratization of AI systems. Nesa introduces a model-agnostic sharding framework designed for decentralized AI inference. Our framework uses blockchain-based sequential deep neural network sharding to distribute computational tasks across a diverse network of nodes based on a personalised heuristic and routing mechanism. This enables efficient distributed training and inference for recent large-scale models even on consumer-grade hardware. We use compression techniques like dynamic blockwise quantization and mixed matrix decomposition to reduce data transfer and memory needs. We also integrate robust security measures, including hardware-based trusted execution environments to ensure data integrity and confidentiality. Evaluating our system across various natural language processing and vision tasks shows that these compression strategies do not compromise model accuracy. Our results highlight the potential to democratize access to cutting-edge AI technologies by enabling secure and efficient inference on a decentralized network.
\end{abstract}

\section{Introduction}
Centralized AI inference poses significant risks related to data privacy, computational bottlenecks, deterministic output, and single points of failure. The prohibitive cost and scarcity of high performance computing resources prevent the mass adoption of the only counter option to centralization, which is open-source decentralized AI. These challenges further limit the ability to contribute to training, fine-tuning, and AI inference at scale. This severely prohibits the adoption of state-of-the-art AI models for enterprises and developers, as well as shared research around the world.

Recent large language models are available with more than 100 billion parameters \cite{jiang2024mixtral,openai2023chatgpt,anthropic2023claude,google2023gemini}, which makes it important to train them on powerful and costly accelerated hardware such as GPUs and TPUs. \cite{muennighoff2024scaling}.
Several approaches can be used to make these models more accessible, especially during the fine-training and inference phase. Using the APIs is one possible approach, which allows quicker inference passes from pre-trained models, but offers little customisation capacity, and no options to change or optimise the training process \cite{openai2023chatgpt,Replicate2023}. A second solution is offloading, where the model components are moved to slower memory (e.g. RAM or SSD) \cite{eliseev2023fast}. Then, only the relevant portion of the model is iteratively moved to the available GPU, allowing it to run on less expensive hardware. However, this process involves frequent data transfers and can be extremely slow for larger models. 

Collaborative efforts to AI execution today are further hindered by the challenges around security~\cite{tovino2016hipaa}. Techniques like continuous and domain adaptation offer partial solutions by enabling model sharing without direct data exchange~\cite{liu2021adapting}. However, these methods are prone to backdoor attacks and often result in sub-optimal model performance due to the limitations of semi-supervised or unsupervised fine-tuning. Collectively, these issues have real-world implications for businesses requiring critical inference. For example,  Traditional centralized systems fail to meet the strict requirements for data privacy in the financial domain, where institutions analyze vast amounts of sensitive transactional data \cite{lee2020autoaudit}. Hence, existing decentralized systems fail to achieve the need for both confidentiality and verifiability.

To address these challenges and enable AI democratization, Nesa introduces the first model-agnostic hybrid sharding approach layered onto a hardware-based and software-based privacy co-optimization protocol. This solution distributes computational loads across multiple nodes on a decentralized inference network. Nesa's distributed inference protocol delivers privacy-preserving data processing while facilitating the scalable and auditable execution of AI inference. The result is a network where anyone can fine-tune models and perform inference queries without substantial investments in computational infrastructure \cite{chen2018machine}. This new protocol, based on its dynamic routing mechanism, greatly lowers the barrier to entry for participating nodes by accommodating various levels of computational capabilities, which is particularly relevant given the high hardware requirements of competing systems that restrict inference participation only to entities with access to top-tier GPUs.

 \begin{figure}[t!]
  \centering
  \includegraphics[width=1.0\textwidth]{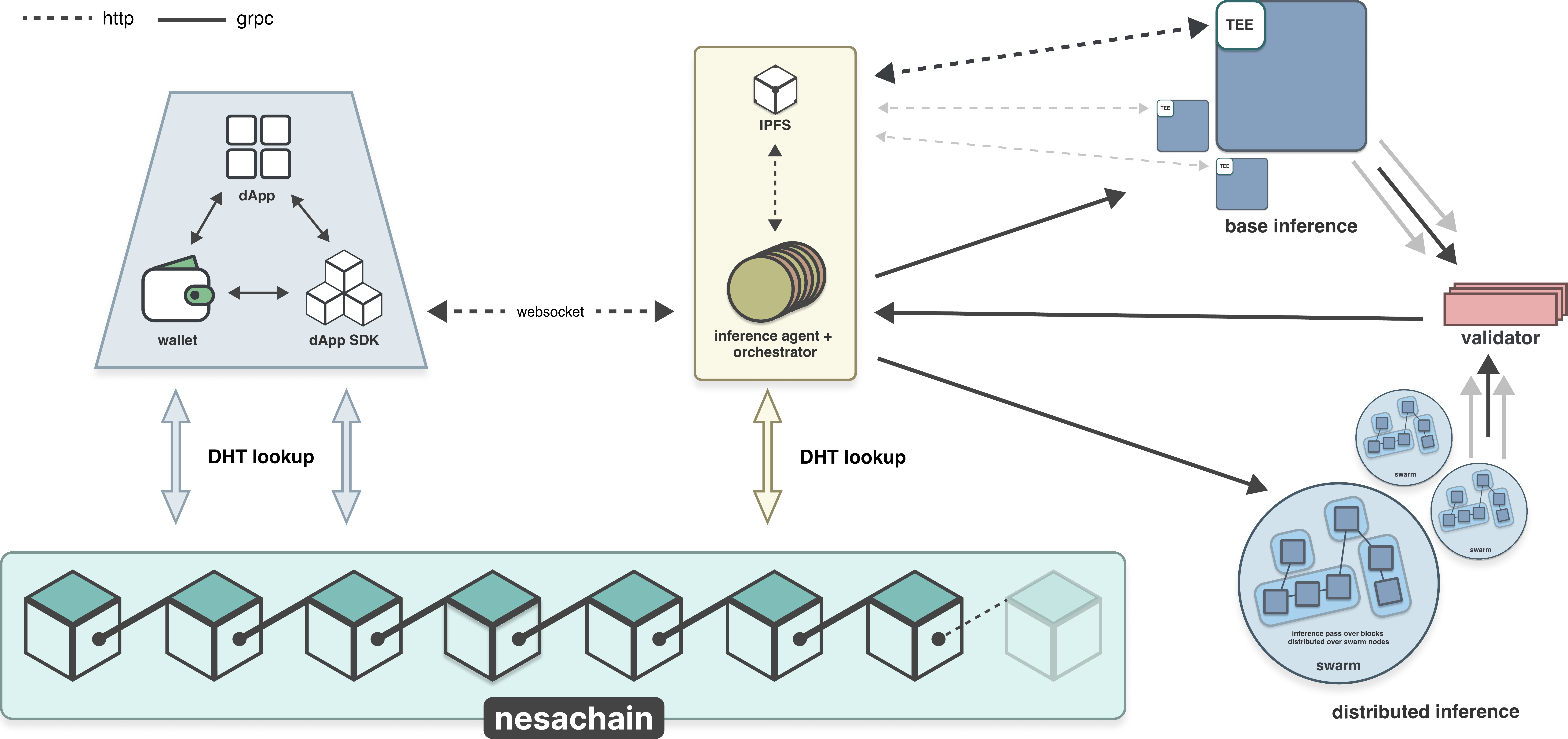}
  \caption{High-level overview of the Nesa network. }
  \label{fig:flowchart}
\end{figure}

We integrate recent model optimization techniques like Zero Redundancy Optimizer (ZeRO), quantization~\cite{jacob2018quantization}, Parameter-Efficient Fine-Tuning (PEFT)~\cite{houlsby2019parameter}, Compression and Low-rank Adaptation (LoRA)~\cite{hu2021lora}, to further improve the computational efficiency of the distributed network and to optimize memory usage and processing capabilities of nodes responsible for serving the model segments. This deployment ensures high model accuracy and throughput across the distributed network, while model shards are secured against tampering via end-to-end encrypted data sharing and aggregation, compliant with regulatory standards for all domains of research and enterprise \cite{touvron2023llama,kirillov2023segment}.

\section{Distributed training and inference}
    Distributed inference and training across multiple nodes are essential due to the exponential increase in the size of models, their complexity and the scarcity of computational resources available to process them \cite{zhang2024scaling}. The essence of both of these tasks lies in the necessity to handle vast computational loads more efficiently and to reduce the latency involved in generating predictions and updating model parameters \cite{moreno2023enhancing}. Our approach uses the collective computational resources and memory available across several processing nodes. This makes it possible to handle larger models or increased batch sizes without a proportional increase in inference or training time. One of the critical aspects of efficient distributed inference and training is partitioning the computational graph of any neural network. This allows the nodes with limited resources to only handle a segment of the model during the inference or training phase. The computational graph represents all the operations and data flows within the model from input to output. By Partitioning this graph we divide the model's computations so that they can be processed in parallel or in sequence across different nodes \cite{lin2023comprehensive,archer2023pipeline}. We integrate optimisation techniques to parallelize the gradient computation and parameter updates which ultimately reduces the speed of the training process. Specifically, we use the ZeRO (Zero Redundancy Optimizer) to split the optimizer state, gradients, and parameters across nodes which reduces the memory footprint and increases the overall scalability. \cite{qi2023zero}.

\subsection{Nesa's model partitioning approach}

Due to the novelty of running deep learning algorithms on the blockchain, Nesa's partitioning mechanism aims to minimize the amount of data that must be transferred between different nodes, thus reducing the latency and bandwidth requirements for both inference and training phases. The overall throughput of the inference and training tasks is limited by the slowest part of the system, known as the bottleneck stage \cite{huang2024re}. We therefore introduce mechanisms to balance the workload and avoid bottlenecks. The partitioning must balance the computational load across all nodes to maximize throughput. This balance ensures that all parts of the system work at their full potential without any single stage becoming a drag on performance.

We prioritize fast memory like Static Random-Access Memory (SRAM) for distributed inference and training across multiple nodes which is essential for storing intermediate model outputs, i.e., activations, and parameter weights \cite{pagliardini2023faster,li2021pipepar}. SRAM is significantly faster than conventional memory solutions but is also more expensive and limited in size. Each node in the network contributes its SRAM, which multiplies the available fast memory during the distributed inference and training session. This increase in capacity allows for caching more model parameters in fast memory.

\subsection{Blockchain-based sequential deep neural network sharding}

Nesa developed a new approach for network topology-informed model sharding, named \textit{Blockchain-based Sequential deep neural network Sharding} (BSNS). Our approach establishes and solves an optimization problem that can find a sequence of nodes able to run an inference request. Each node will be typically a distributed machine on the network, which will normally run a block or a contiguous sequence of blocks of a deep learning architecture. It is crucial that for each inference session involving block sharding, a chain of nodes is found that can collectively reconstruct the layers of the full model \cite{yang2022blockchain}. 

The key innovation of our approach is that the selection of nodes for sharding is informed by: (i) the topology of the network; (ii) our persistent homology metrics based on graph embeddings and (iii) network-based variables, including latency and geographical distance between the nodes. Taken together, this will constitute a full end-to-end neural network that performs inference across the blockchain at the optimal speed, fully embedding the topology and the security components of the blockchain.

\subsubsection{Swarm creation and dynamic rebalancing}
\label{sec:swarm_creation}
The BSNS framework allows arbitrary deep neural networks to be distributed for training, but with specific focus given to LLMs and transformer-based architectures. The swarm is formed based on a heuristic that selects the optimal set from the pool of available nodes. Each node inside the swarm processes a shard of the network architecture and communicates through a sequence of remote procedure calls (RPC). Each $N_i$ denotes a node within the network which is responsible for the $i$-th shard of the model. Whereas, $p$ represents the total number of shards distributed across the swarm. The selection process heuristic considers different network topology features, such as node and edge structure, bandwidth, and computational capabilities, therefore ensuring optimal efficiency in data processing across the swarm. This has strong practical benefits for our clients, as distance and latency heavily affect the performance when computing using a blockchain.

Given a network $A$ with $n$ nodes that need to execute $p \leq n$ shards of a model, and assuming that each block is held by one network node, the task involves finding a sequence $\mathcal{S}$ of nodes
\begin{equation}
\mathcal{S}: (A_i)_{i\in{1,\ldots,p}}. 
\label{eq:A}
\end{equation}

We model this as a recursive sequence finding problem. Hence, the selection of each node depends on the previous node(s) shortlisted based on the selection heuristic parameters.
\begin{equation}
A(i) = A_i = f((A_{i-1-\alpha},\ldots,A_{i-1}); \ \text{network parameters}(\alpha)), 
\label{eq:A_function}
\end{equation}

 \begin{figure}[t!]
  \centering
  \includegraphics[width=0.85\textwidth]{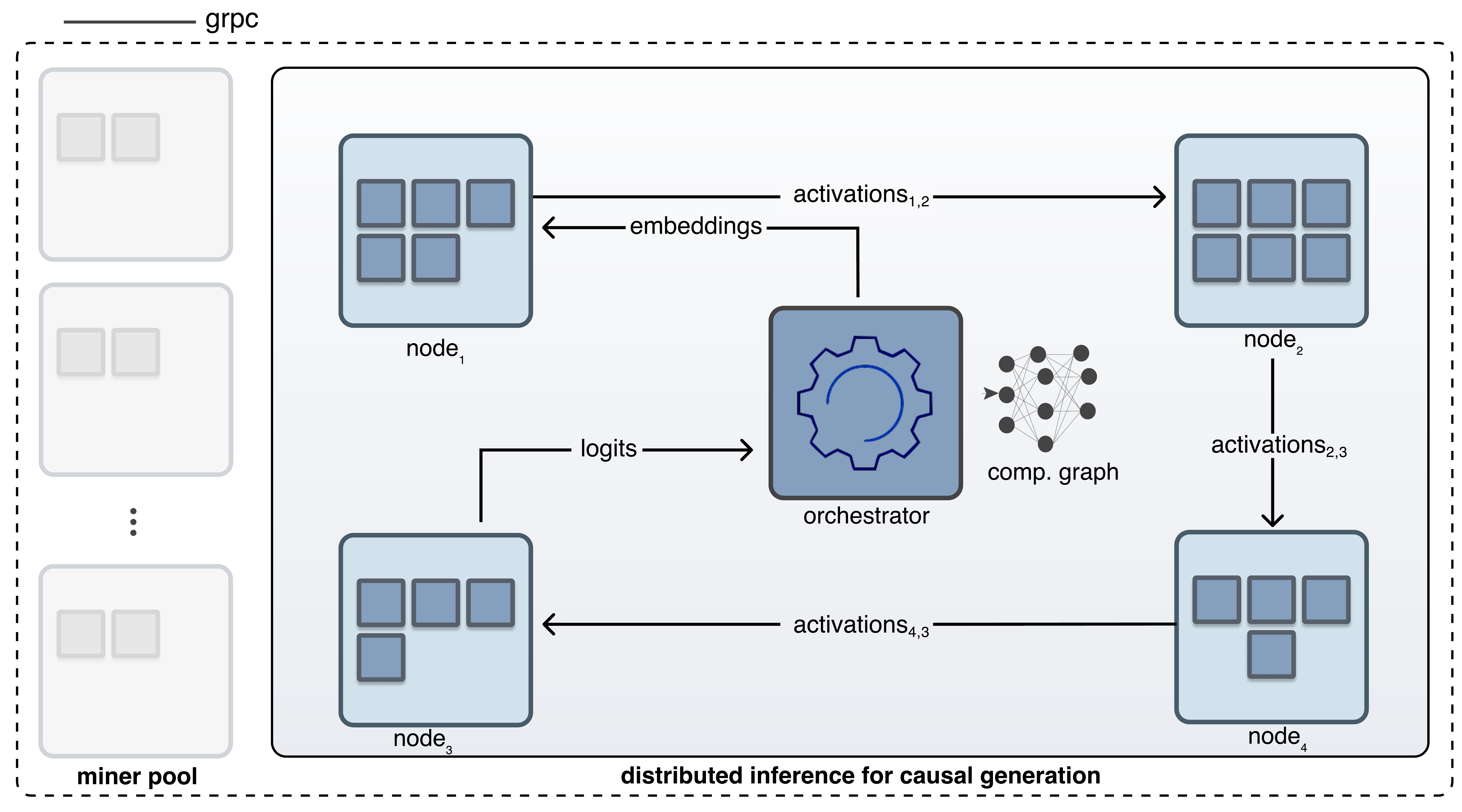}
  \caption{Orchestrator is chosen from the pool of accessible nodes and picks nodes to build a whole set of blocks (e.g., transformer blocks) according to the model architecture that needs to be run. To enable direct routing between blocks, each node has a unique section of the computational graph. Initiating from the orchestrator, the inference is routed to the first block and then through each remaining block directly in order.}
  \label{fig:distributed_inference}
\end{figure}

where $\alpha < i-1$ sets the context length for network inference and affects the computation of the network parameters. If $\alpha=0$, the network parameters only involve distance metrics between the current node and the previous node, while for $\alpha>0$, more than one node is considered.  The BSNS framework introduces a dynamic rebalancing algorithm that adjusts based on node performance and adapts to adverse network conditions. This algorithm monitors the network state and status which allows redistribution of model shards among nodes over time. The swarm rebalance is given by:

\begin{equation}
{R : S \times P \rightarrow S'}
\end{equation}

where \( R \) represents the rebalancing function that transforms the current sequence of nodes \( S \) based on the network parameters \( P \). Like the selection heuristic, this function considers features such as hardware capacity, throughput, and bandwidth, which influence data transmission speeds between nodes.

\subsection{Cache optimization to enhance efficiency}
BSNS uses a Key Value (KV) cache for scaling LLMs across a distributed network. This minimizes the computational overhead associated with token generation in LLMs when dealing with models that operate on a token-by-token basis \cite{dong2024get,jiang2023mistral}. This mechanism uses the inherent characteristics of transformer models by caching the key and value vectors after their initial computation to prevent redundant calculations in subsequent generations. This also decreases the load on the network's nodes and allows the processing of longer sequence lengths in LLMs \cite{dong2024qaq}.

Caching improves efficiency by ensuring that each node within a swarm can generate tokens by utilizing pre-computed key and value vectors. This is a core factor that increases the overall throughput of the system. Moreover, it enables the system to scale to models with extensive context sizes by removing the memory and computational constraints that typically hinder the operation of large LLMs.

\begin{equation}
    \text{Cache Size per Token} = 2 \cdot \text{head\_dim} \cdot n\_heads \cdot n\_layers, 
    \label{eq:cache_size}
\end{equation}
where \textit{$\text{head\_dim}$} represents the dimensionality of the key and value vectors,\textit{ $n\_heads$} is the number of attention heads, and $n\_layers$ is the number of transformer layers within the model. This highlights the direct relationship between the model's complexity and the size of the KV cache which shows the need for effective cache management strategies \cite{zhao2024alisa}.
\subsection{BSNS with parameter-efficient fine-tuning via adapters}

BSNS uses Low-rank adaption techniques to enable fine-tuning capabilities for arbitrary language models \cite{hu2021lora}. The adapters are inserted between the transformer layers which allows learning the downstream tasks without retraining the entire network as it can be computationally expensive\cite{hu2021lora,yu2024investigating}.

\subsubsection{Operational framework for adapters in Nesa}
Adapters are small neural network modules inserted between the layers of a pre-trained model. They allow task-specific training with minimal parameter updates \cite{hu2021lora}. The operation of an adapter within a transformer layer is given as:
\begin{equation}
h'_i = \text{LayerNorm}(h_i + W_2 \phi(W_1 h_i)),
\end{equation}
where \(h_i\) is the input to the adapter at layer \(i\), \(W_1\) and \(W_2\) are trainable weights of the adapter, \(\phi\) represents a non-linear activation function, and \(h'_i\) is the output of the adapter \cite{dettmers2024qlora}. 

The adaptation of a weight matrix \(W\) in a transformer can be modeled as
\begin{equation}
W' = W + BA,
\end{equation}
where \(W\) is the original weight matrix of the model, \(A \in \mathbb{R}^{r \times n}\) and \(B \in \mathbb{R}^{n \times r}\) are the low-rank matrices introduced by LoRA with \(r \ll n\), and \(W'\) is the adapted weight matrix \cite{hu2021lora}. Upon initiating a fine-tuning session, each participating node initializes adapter modules according to the specified configuration aligned with the shard of the LLM it is responsible for.

\subsubsection{Node synchronization}

Nodes collaboratively fine-tune their adapter modules using gradients from task-specific data processed through their portion of the LLM. The process includes three main steps:

1. \textit{Forward pass.} Nodes perform a forward pass through transformer layers and adapters which generates activations based on the input tensors.

2. \textit{Backward pass.} From the final layer of a large language model, gradients are propagated by comparing predictions with true labels. Gradients for corresponding adapters are used to perform the gradient descent update.

3. \textit{Synchronization and update.} Nodes synchronize their updates across the network to ensure consistency and convergence

The parameter updates for the adapters, denoted as \(\theta_i\) for the \(i\)-th node, follow the rule:

\begin{equation}
\theta^{(t+1)}_i = \theta^{(t)}_i - \eta\nabla_{\theta_i} L(\theta^{(t)}_i)
\end{equation}

where \(t\) is the current iteration, \(\eta\) is the learning rate, and \(L\) is the loss function.

Along with these synchronized update rules, BSNS uses a consensus mechanism for agreeing on fine-tuning objectives, data distribution, and synchronization intervals. This setup allows for the sharing and reuse of fine-tuned adapters and allows nodes to build on existing adaptations for new tasks.



\subsection{Dynamic sharding of neural networks}
The BSNS method works well for LLMs as they have common blocks that can be distributed on a sequence of nodes. However, for arbitrary layered neural networks like diffusion or sequence models \cite{croitoru2023diffusion,grossberg2013recurrent}, this is not possible because of wide architectures and multimodal prediction heads. Nesa utilizes a dynamic sharding scheme to optimally partition the network's computation graph, $G=(V, E)$, across multiple nodes. This graph comprises operations $V$ and data flow edges $E$. Each operation $v \in V$ outputs a tensor, forming an edge $(u,v) \in E$. It includes all computational tasks from basic arithmetic to layer-specific matrix multiplications, each with unique computational and memory requirements: execution time $\text{work}(v)$, parameter memory $\text{sizeparam}(v)$, and output size $\text{sizeout}(v)$ \cite{qi2023zero}. 

Partitioning this graph involves dividing $V$ into $k$ distinct blocks such that each block can be processed on a different node in a swarm under the constraint that the induced quotient graph of $G$ remains acyclic. This division aims to maximize throughput while minimizing inter-node communication subjected to the bandwidth $B$ between nodes, with the I/O cost from node $S$ to node $T$ given by: 

\begin{equation}
\text{io}(S,T) = \frac{1}{B} \sum_{v \in N^-(T) \cap S} \text{sizeout}(v), 
\label{eq:io}
\end{equation}
where $N^-(T)$ represents the set of nodes whose outputs are consumed by block $T$.

The main challenge is effectively distributing the model's parameters and activations across each node's fast memory (e.g., SRAM) \cite{saha2024complexity}. Parameters that exceed the fast memory capacity are streamed from slower storage and add latency. The overflow cost, representing the time to stream parameters beyond the fast memory limit $M$, is calculated as:

\begin{equation}
\text{overflow}(S) = \left(\text{sizeparam}(S) + \text{peak}(S) - M\right) + \frac{\text{peak}(S)}{B}, \label{eq:overflow}
\end{equation}
where $\text{peak}(S)$ denotes the peak memory requirement for activations within block $S$.

The block cost, $f(S)$, combines the costs of receiving input tensors, executing the block's operations (including any overflow cost from streaming parameters), and sending output tensors downstream:

\begin{equation}
f(S) = \text{io}(V\setminus S,S) + \sum_{v \in S} \text{work}(v) + \text{overflow}(S) + \text{io}(S,V\setminus S). \label{eq:f_function}
\end{equation}

The goal of this partitioning, defined by the Max-Throughput Partitioning Problem (MTPP) \cite{archer2023pipeline}, is to minimize the maximum cost across all blocks, optimizing the throughput of the entire pipeline. Formally, MTPP seeks a partition $P^*$ that minimizes the bottleneck cost:
\begin{equation}
P^* = \text{argmin}_{P \in P_k(G)} \left\{ \max_{i\in[k]} f(P_i) \right\}, 
\label{eq:P_star}
\end{equation}
where $P_k(G)$ denotes the set of all possible partitions of $G$ into $k$ blocks, and $\text{cost}^*$ is the minimum achievable bottleneck cost across these partitions.

\subsection{Enhanced MTPP slicing of topological order}
 The MTPP algorithm is key in the optimization of computation graphs for neural network inference across multiple systems. It is known to be NP-hard, as shown by its relation to the minimum makespan scheduling problem \cite{tremblet2024makespan}. This makes fully polynomial approximations unlikely. Instead, we use a heuristic approach to handle this complexity and maintain high throughput. For non-transformers based architectures, our method combines Kahn's algorithm \cite{liu2014low} for topological sorting with dynamic programming and segment trees. Kahn's algorithm ensures correct topological order for acyclic partitioning. Dynamic programming then calculates the best partitioning and Segment trees quickly adjust partitions during optimization. This combination helps in distributing the computational load evenly and reduce communication delays \cite{xia2023end, yamagata2023q}. 

Our approach centres around the \textit{segment cost data structure} which calculates costs based on computation and communication for each segment. We start with a computation graph $G$ and a topological order $\pi$. The \texttt{SliceGraph} algorithm uses this to divide the graph into throughput-optimized segments. It dynamically splits the graph into blocks which distributes workload evenly and reduces communication between nodes. 

\begin{algorithm}[t!]
\caption{Enhanced MTPP Slicing of Topological Order $\pi$ into at most $k$ Blocks}
\begin{algorithmic}[1]
\Function{EnhancedSliceGraph}{$G, k, \pi$}
    \State Initialize \textit{enhanced\_segment\_cost} data structure for $(G, \pi)$ using heuristic-based weighting
    \State \Return \Call{EnhancedDP}{\textit{enhanced\_segment\_cost}, $n$, $k$}
\EndFunction
\Statex
\Function{EnhancedDP}{\textit{enhanced\_segment\_cost}, $r$, $k'$}
    \If{$k' = 1$}
        \State \Return \textit{enhanced\_segment\_cost.Query}$(1, r)$ \textit{// Leverage the enhanced query for improved cost estimation}
    \EndIf
    \State $ans \gets \infty$
    \For{$\ell = 1$ \textbf{to} $r - 1$} \textit{// Iterate through the nodes}
        \State $a \gets$ \Call{EnhancedDP}{\textit{enhanced\_segment\_cost}, $\ell$, $k' - 1$}
        \State $b \gets$ \textit{enhanced\_segment\_cost.Query}$(\ell + 1, r)$
        \State $ans \gets \min(ans, \textit{heuristic\_adjustment}(\max(a, b)))$
    \EndFor
    \State \Return $ans$
\EndFunction
\end{algorithmic}
\end{algorithm}

The initialization of the segment cost data structure for a graph with $n$ nodes is $O(n^2)$ which allows us to pre-compute all possible segment costs. This computation is important for the correct operation of the SliceGraph algorithm \cite{barros2012assertion}. Initially, the algorithm's time complexity is $O(n^2k + m\log^2(n))$, where $k$ is the number of partitions, and $m$ is the edge count. This includes the dynamic programming process and cost queries. Applying the convex hull trick reduces the complexity to $O(nk\log n)$ and speeds up optimal partitioning.

\subsubsection{Biased random-key genetic algorithm}
BRKGA improves our approach by using stochastic and evolutionary methods which overcomes the limitations of heuristic-based techniques \cite{sivanandam2008genetic,londe2024biased}. This algorithm enables broad exploration of partition configurations, essential for solving the NP-hard MTPP. Over generations, BRKGA evolves partitioning strategies to discover solutions that balance computational load and reduce communication overhead and achieves optimal throughput in distributed deep neural network inference \cite{archer2023pipeline}.


BRKGA evolves node priority vectors \(\mathbf{x} \in [0,1]^n\) and links these priorities directly to partitioning quality for adjustments. The process starts by creating a population of random vectors. These vectors help setup orders in the SliceGraph algorithm. Each vector's quality is assessed for throughput optimization and latency reduction which determines its fitness. BRKGA refines the strategies until convergence criteria for node priorities is met. This process involves selection, crossover \(\mathit{new} = \alpha \mathit{parent}_1 + (1 - \alpha) \mathit{parent}_2\), and mutation to generate near optimal offsprings.

The BRKGA approach offers many advantages over sequential sharding methods for partitioning neural networks. Sequential methods follow a predetermined path which perhaps can miss more efficient configurations that could improve computational throughput and reduce communication latency. This adaptability is important in distributed environments with varying network topologies and node capabilities. 

\subsection{Swarm topology}

Due to the dynamic topology of the swarm network, we further extend our computational efficiency by monitoring and classifying changes and the current state of our swarm topology over time.  In conjunction, we build several pre-computed sharding schemas based on the commonly observed swarm topologies as we monitor our networks. In doing so, we can quickly select among sharding schemes to find one that is near-optimal for the current swarm topology.  In the following section, we outline how we monitor and analyze the swarm topology.

\subsubsection{Topological forecasting and monitoring}
\label{Topological}
Dynamic sharding is computationally expensive as finding an optimal sharding scheme is known to be NP-hard \cite{archer2023pipeline}. We mitigate this expense by precomputing highly efficient sharding schemas for different topological states of our network on a per-model basis.  As the network topology itself is dynamic, we periodically monitor our network using techniques from (weighted) persistent homology \cite{meng2020weighted,goerss2009simplicial}, to give snapshots of topological characteristics to guide the applications of our pre-computed sharding schemas.  The main goal of this section will be to introduce many of the concepts behind persistent homology (which is not a mainstream topic) to guide intuition on how this computational tool can be leveraged to make dynamic sharding optimization much more tractable. 

As a basic introduction to the implementation of this framework, we briefly illustrate the mathematical technique of simplicial homology. This uses the simplicial complex to calculate the homology groups -- these groups are formally Abelian groups but they can be thought of as vector spaces.  A simplicial complex is a collection of simplices that are the convex hull of a group of vertices.  More explicitly, an  $n$-simplex $\Delta^n $ contains $n+1$ vertices $\{v_0, v_1, \dots, v_n\}$ and is the collection of points in $\mathbb{R}^n$ such that:
\begin{equation}
\Delta^n = \left\{k_0v_0 + \cdots + k_nv_n \ \vert\  \sum_{i=0}^n k_i =1 \textrm{ and } k_i \geq 0 \textrm{ for } 0\leq i\leq n \right\}.
\end{equation}

Thus, a 0-simplex $\Delta^0$ is a point, a 1-simplex $\Delta^1$ is a line segment, a 2-simplex $\Delta^2$ is a triangle, and so on.  To build a simplicial complex out of a collection of simplices, we need to define how to glue the pieces together \cite{albertsson2020simplicial}.  Mathematically it is given as (linear) boundary maps $\partial_n^i: C^n \to C^{n-1}$ which map each $n$-simplex $\Delta^n_i$ to its boundary which lives in one dimension lower (if it exists).  Here, were are using the notation $C^n$ to indicate the collection of all $n$-simplices in our complex, sums of which are often referred to as $n$-chains or simply chains.  In general, we can (abstractly) represent the entire data of the simplicial complex by:
\begin{equation}
 \cdots  \xrightarrow[]{\partial_{n+2}} C^{n+1} \xrightarrow[]{\partial_{n+1}} C^n \xrightarrow[]{\partial_{n}} C^{n-1} \xrightarrow[]{\partial_{n-1}}\cdots
\end{equation}

\begin{figure}[t!]
  \centering
  \includegraphics[width=1.0\textwidth]{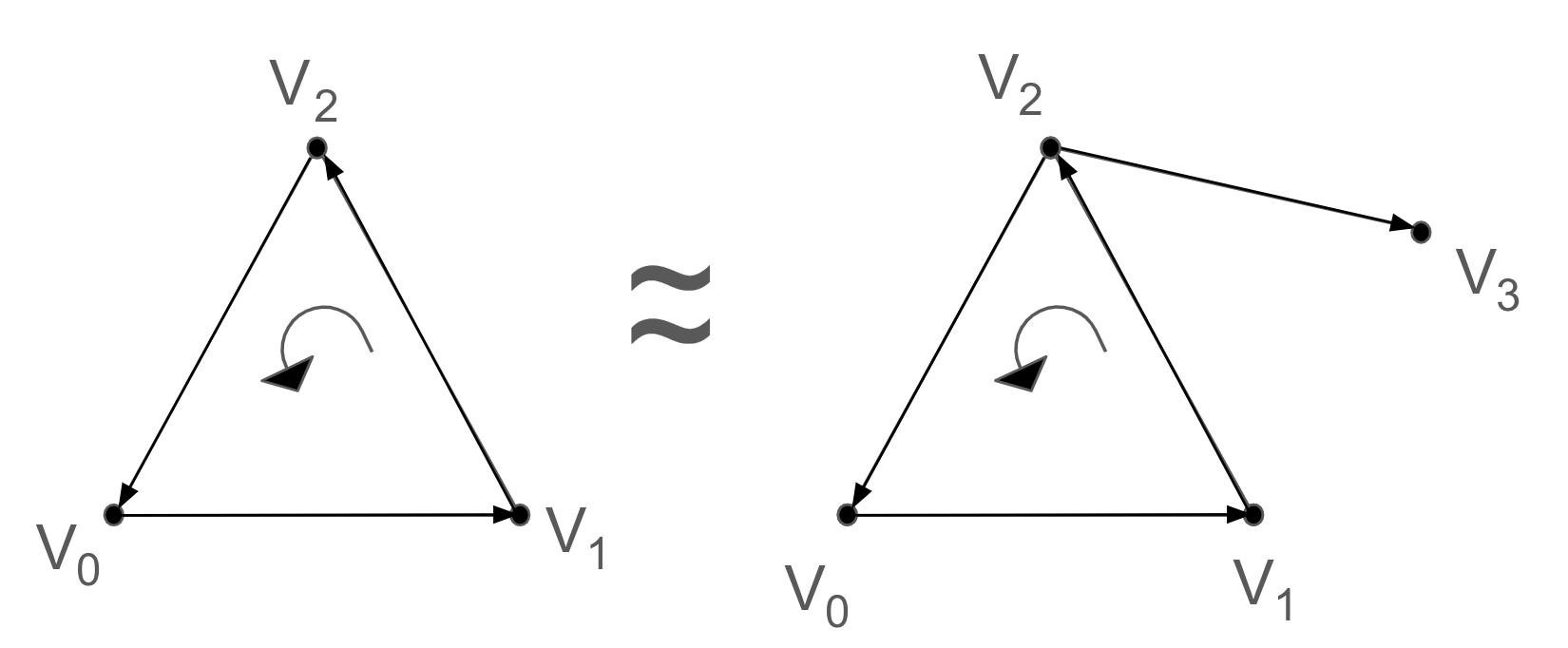}
  \caption{Toy example for simplicial homology: the left simplicial complex consists of three vertices (0-simplices) and three edges (1-simplices) whereas the right consists of four vertices and four edges.  Topologically, these two complexes are identical from the viewpoint of simplicial homology, but they are distinguishable via persistent homology.}
  \label{fig:simplical_homology}
\end{figure}

Referring back to the boundary maps  $\partial_n^i: C^n \to C^{n-1}$, the image of $\partial_n^i(\Delta^n_i)$ will be the (simplicial) boundary of $\Delta^n_i$. We assign an orientation to the simplices to make boundary calculations consistent. For example, with the 1-simple as the convex hull of $\{v_0, v_1\}$ the image of the boundary map would be defined as $v_1 - v_0$, with such an orientation induced by imaging the 1-simple is a directed edge from the point $v_0$ to the point $v_1$.  More explicitly:
\begin{equation}
\partial_1(\Delta_0^1) = v_1 - v_0,
\end{equation}
where we have intentionally denoted $\Delta_0^1$ as the simplex constructed from the vertices $\{v_0, v_1\}$.  Looking at Figure \ref{fig:simplical_homology}, and considering the triangle at the left, the collection of 1-chains contains three edges, which are the convex hulls of the sets, $\{v_0, v_1\}$, $\{v_1, v_2\}$ and $\{v_0, v_2\}$.  The arrows give the orientation of each edge, which guides the boundary map calculation.  We already calculated the boundary map for the first edge.  The remaining two edges, $\Delta_1^1$ and $\Delta_2^1$ map can be calculated as:
\begin{equation}
\begin{aligned}
\partial_1(\Delta_1^1) &=v_2 - v_1\\
\partial_1(\Delta_2^1) &= v_0 - v_2
\end{aligned} 
\end{equation}

By linearity, we can compute the boundary for the full chain, $\Delta_0^1 + \Delta_1^1 + \Delta_2^1$ as:

\begin{equation}
\begin{aligned}
\partial_1(\Delta_0^1 + \Delta_1^1 + \Delta_2^1) &= \partial_1(\Delta_0^1) + 
\partial_1(\Delta_1^1) + \partial_1(\Delta_2^1)\\
&= (v_1 - v_0) + (v_2 - v_1) + (v_0 - v_2)\\
&= 0.
\end{aligned}
\end{equation}

Thus, the chain $\Delta_0^1 + \Delta_1^1 + \Delta_2^1$ has no formal boundary (to be more clear this chain is what most people identify as the triangle itself).  More importantly. This example demonstrates that a cycle (a chain without a boundary) has a trivial image.  This will correlate with some topological holes in the simplicial complex.  Thus, cycles correspond to elements of the kernel of the boundary map (a kernel of a map is simply all elements that map to zero).  Formally, we calculate the $n$th-homology group $H_n(C^\star)$ of a simplicial complex $C^{\star}$ as 
\begin{equation}
H_n(C^\star) = \frac{Z_n(C^\star)}{B_n(C^\star)},
\end{equation}
where $Z_n(C^\star)$ is the group of $n$-chains that are cycles and $B_n(C^\star)$ are the group of $n$-chains that are boundaries. 

Thinking of these groups as vector spaces with each dimension corresponding to a simplex and with the operation being vector addition, this definition allows one to identify chains where the difference is a boundary (a chain that has a non-trivial image). In other words, adding a boundary chain to a cycle does not change its class under homology.  This is demonstrated in Figure \ref{fig:simplical_homology} where the two diagrams are equivalent since the difference in $C^1$ is the edge between $\{v_2, v_3\}$. Note that this is a boundary because the value under the boundary map is $v_3 - v_2 \neq 0$, according to the indicated orientation.  It should be further noted that points never have boundaries so $H_0(C^\star)$ does not count holes, but counts the number of connected components. Overall, this mathematical formalism allows one to unambiguously tabulate the topological characteristics of a wide variety of different geometric objects.

Persistent homology \cite{carlsson2007theory} (PH) is based on introducing a filtration of the topological space as a sequence of simplicial complexes. For our purposes, the complexes will be derived from the graph corresponding to our network.  For clarity, we will refer to the persistent $n$th homology group for a space $M$ as $PH_n(M)$. For our application, the embedding of this graph into Euclidean space is based on the geographic locations and the edge weighting is a function of individual node internet connectivity and node computational power. 

 \begin{figure}[t!]
  \centering
  \includegraphics[width=1.0\textwidth]{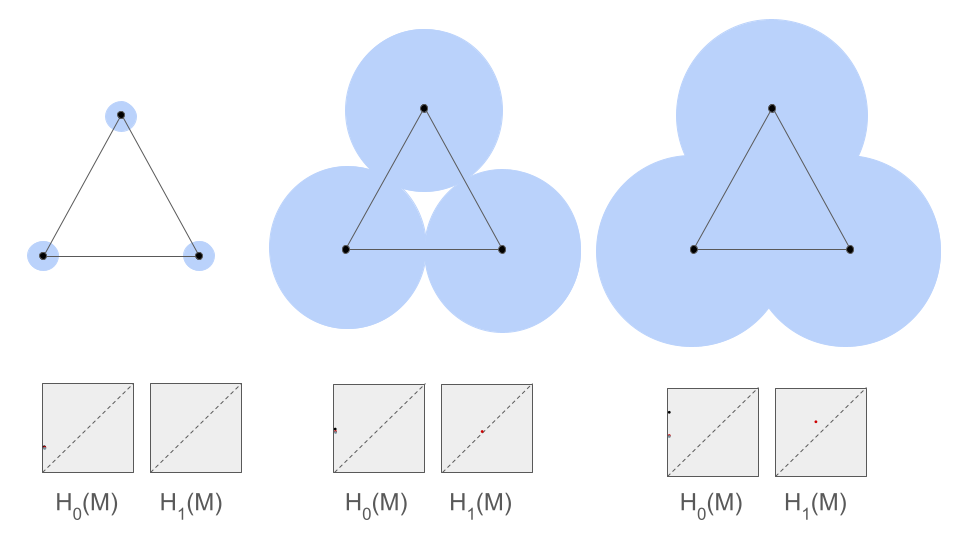}
  \caption{Visualizing persistent homology for a simple graph of 3 vertices on an equilateral triangle: the top row illustrates 3 snapshots of the filtration sequence of three circles of growing radii (filtration parameter).  The bottom row demonstrates the respective persistence diagrams.  Note that in $H_0(M)$, the initial 3 dots are shifted slightly to aid the eye but they would overlap precisely in reality.}
  \label{fig:ph_diagram}
\end{figure}

The filtering process in persistent homology for a graph resembles inflating circles around each network node (vertex) and capturing data about how the resulting topology changes as the circles begin to merge.  One can gain intuition behind this idea by imagining three points on the vertices of a triangle (see Figure \ref{fig:ph_diagram}).  A scale parameter increases the radius of each point uniformly. Each step along this parameter change introduces a new step in the filtration of complexes.  Until the circles grow enough to overlap, there will be three separate components so the zeroth homology group which measures the number of connected components will have dimension 3 and higher order groups which measure the number of higher dimensional holes will be trivial (left schematic in Figure \ref{fig:ph_diagram}).  As the radii grow they eventually grow to overlapping circles.  Immediately after the circles overlap there will still be a gap between the three circles centered in the middle of the triangle, so now they’re one hole and thus the first homology group suddenly becomes non-trivial with dimension 1 (center schematic in Figure \ref{fig:ph_diagram}).  At the same time, the number of connected components collapses to 1 so the zeroth homology group will change from three dimensional to 1 dimensional.  As the circles continue to grow, the gap in the middle will fill, closing the hole and the first homology group will become trivial again, but the zeroth homology group will continue to have dimension 1 (right schematic in Figure \ref{fig:ph_diagram}).   No further meaningful changes will occur to the persistent diagram after this point.   

The changes illustrated above are typically captured using a persistence diagram which simply introduces a point for each birth-death pair, where the x-axis is the birth time and the y-axis is the death time. Here time is just the scaling parameter.  Since births always occur before deaths, all the points on the diagram will be above the main diagonal line $y=x$.  Following the toy example above, one can imagine that smaller holes will have quicker deaths since they do not persist long, so they will be near the main diagonal.  Conversely, bigger holes will have a longer lifetime.  Intuitively, this gives information about the network latency, connectivity and possible bottlenecks.   The lifetimes of the connected components, measured by $PH_0(M)$, give a complementary view of these characteristics.  This data from both $PH_0(M)$ and $PH_1(M)$ persistence can be fed into a dedicated ML classifier to predict which precomputed sharding scheme will be most efficient based on the current network topology.  As calculating persistent homology is not NP-hard \cite{boissonnat2018computing} even in dimensions much higher than those used to model graphs, this gives us an avenue to select a near-optimal precomputed sharding scheme at a reasonable computational cost.

\section{Homology enabled routing mechanism}

Nesa's inference framework is based on a Distributed Hash Table (DHT) for routing data and tasks on the distributed blockchain network \cite{dabek2005distributed}. This is a critical factor in task allocation, and fault tolerance as well as ensuring node information can be accessed globally. The DHT is immediately updated when a node becomes unreliable to change and update the heuristic parameters that allow the network the ability to redeploy without intervention. In the BSNS framework, this becomes even more important as tasks are distributed across a swarm of nodes.

The effective routing of data and tasks is crucial in a distributed blockchain network designed for neural network computations. The routing decision specifies a subset of nodes that are chosen to serve as the computational resources and, thus influence system-wide capacity efficiency, latency and reliability. Nesa's routing mechanism is dynamically updated based on the node features and is given as:
 \begin{figure}[t!]
  \centering
  \includegraphics[width=0.8\textwidth]{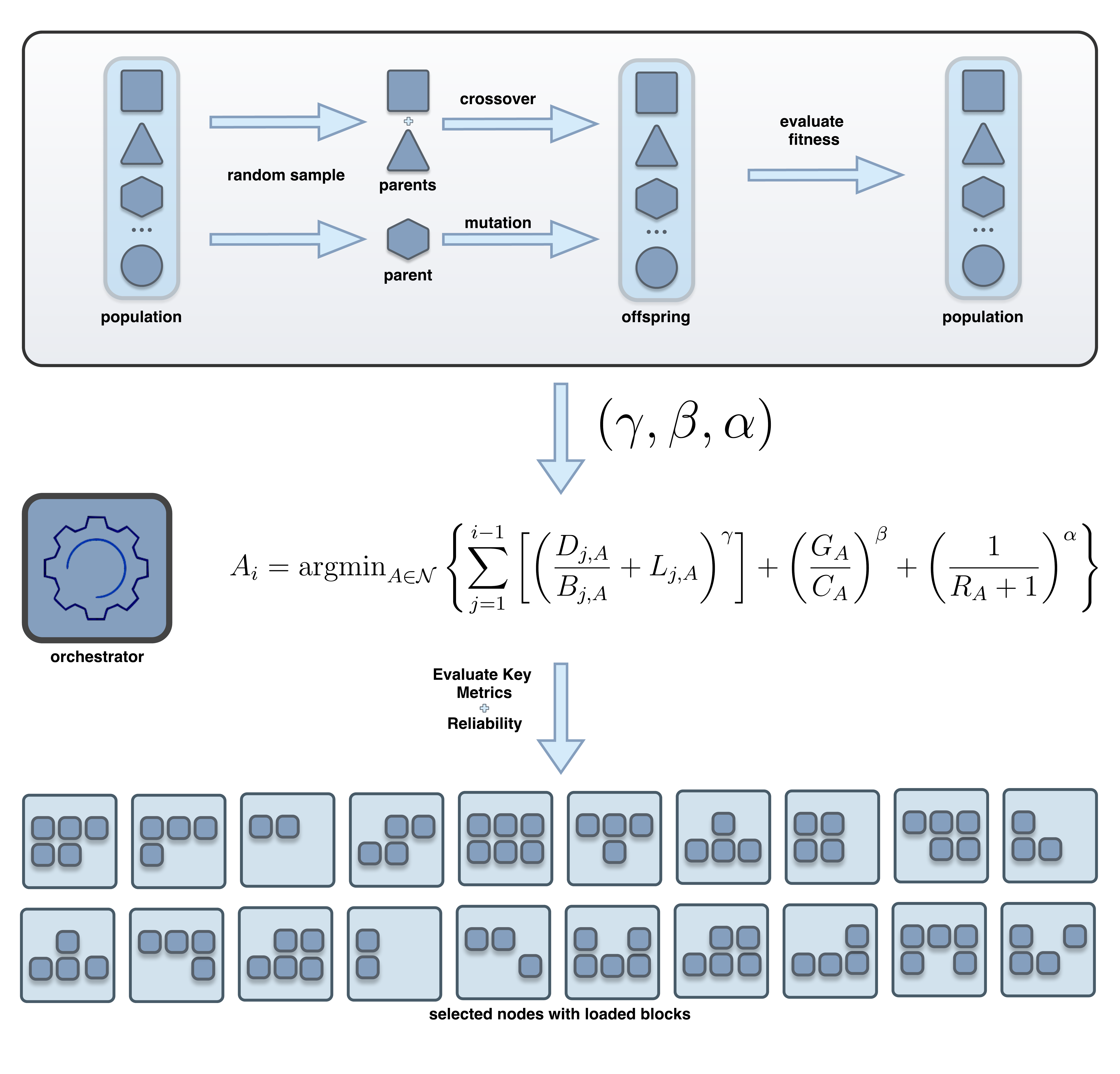}
  \caption{Overview of genetic algorithms for optimization and a decision-making framework based on multi-objective optimization metrics for selecting nodes in a network.}
  \label{fig:genetic_algo}
\end{figure}
\begin{equation}
A_i = \text{argmin}_{A \in \mathcal{N}} \left\{ \sum_{j=1}^{i-1} \left[ \left(\frac{D_{j,A}}{B_{j,A}} + L_{j,A}\right)^{\gamma} \right] + \left(\frac{G_A}{C_A}\right)^{\beta} + \left(\frac{1}{R_A + 1}\right)^{\alpha} \right\}
\end{equation}

We use multiple key metrics which consider the data processing factors, computational load ratio and reliability of the nodes for forming the swarm and its dynamic rebalancing. The key data processing factors are as follows: 

1. \textit{Data Size (\( D_{j,A} \)).} It measures the size of the payload that needs to be transferred from node \( j \) to node \( A \). This factor affects multiple components including model parameters, activations of intermediate layers and final outputs. 

2. \textit{Bandwidth (\( B_{j,A} \)).} This metric measures the maximum data transfer capacity between two nodes per unit of time. Better bandwidth indicates a higher capacity for handling large data volumes and reduces the likelihood of bottlenecks. 

3. \textit{Latency (\( L_{j,A} \)).} Latency is the delay experienced in communication between two nodes. It can be influenced by physical distances and network congestion. Lower latency is important in enabling faster fine-tuning and inference. 

These factors measure the combined cost of data transfer and latency. The hyperparameter \(\gamma\) is used for non-linear scaling of this cost and weights the impact of these factors on overall efficiency. The parameter \(\gamma\) is tuned using genetic algorithms and represents the relative importance of minimizing data transfer time and latency in different scenarios.

Along with these data processing factors we use computational load ratio. This term expresses the balance between available hardware resources and the current computational load. A higher ratio indicates that a node has more resources available relative to its load which makes it a better candidate for selection in a swarm.

1. \textit{GPU Availability (\( G_A \)).} This metric represents the available GPU resources at a node. Nodes report their total GPU capacity and current usage calculated as follows:
    \[
    G_A = \text{total GPU capacity} - \text{current GPU usage}
    \]
This helps identify nodes with sufficient GPU resources to handle inference and fine-tuning tasks without significant delays.
    
2. \textit{Computational Load (\( C_A \)).} This measures the current computational burden on a node, including CPU and memory usage. It reflects the node's active tasks and overall workload. Monitoring \( C_A \) is important to prevent overburden and inference failures.

 For computational load ratio, the parameter \(\beta\) adjusts the sensitivity of this ratio, prioritizing nodes with sufficient GPU resources for compute-intensive models. We also consider a resource reputation metric (\( R_A \)) that assesses the reliability of a node. It incorporates factors such as historical uptime, failure rates, and maintenance history. A node with high uptime and low failure rates is considered more reliable. The calculation \( R_A = 1 - \text{Uptime} \) provides a simple measure of reliability, where uptime is reported as a percentage. The parameter \(\alpha\) controls the weight of reliability in the overall cost calculation, with higher values increasing the importance of selecting reliable nodes.

\subsection{Persistent homology integration}

Persistent homology in our routing mechanism provides insights into the swarm's structure including the stability and connectivity of nodes. It analyzes the persistence of topological features like connected components and cycles, which helps to identify stable regions in the network important for the node selection process. Routing uses persistent homology for network analysis, filtration and node selection. A filtration is constructed by adding simplices in a specific order. This process shows how different features of the network emerge or disappear as the scale changes. For example, at a small scale, only the strongest connections (shortest edges) are included as they represent the core structure of the network. As the scale increases, weaker connections are added. 

Nodes that are part of long-lived features are considered more stable and reliable. For example, a node consistently appearing in a connected component with a high $\beta_0$ value across multiple scales will likely reside in a stable network region. This stability influences the adjustment of $R_A$ values in the routing mechanism and favors nodes in these regions. If persistent homology shows that a cluster of nodes consistently forms a connected component, these nodes are deemed more reliable, impacting their $R_A$ values. 

\subsection{Evolutionary algorithm for parameter tuning}

The parameters \(\gamma\), \(\beta\), and \(\alpha\) are critical in balancing the cost components in the routing mechanism. We use a genetic algorithm \cite{lambora2019genetic} to optimize these parameters which allows routing to adapt to varying network conditions and workloads.

1. \textit{Initialization.} The algorithm starts with a population of candidate parameter sets, each represented by a tuple \((\gamma, \beta, \alpha)\). These candidates are initialized randomly in a predefined range for a diverse starting point for the optimization process.

2. \textit{Fitness Evaluation.} The fitness of each candidate is evaluated based on the total cost \( F(\gamma, \beta, \alpha) \). The fitness function is inversely related to the cost where lower costs indicate higher fitness. This step involves calculating the objective function for each candidate, using real-time data from the network.

\begin{equation}
F(\gamma, \beta, \alpha) = \sum_{i=1}^{n} \left\{ \sum_{j=1}^{i-1} \left[ \left(\frac{D_{j,A}}{B_{j,A}} + L_{j,A}\right)^{\gamma} \right] + \left(\frac{G_A}{C_A}\right)^{\beta} + \left(\frac{1}{R_A + 1}\right)^{\alpha} \right\}
\end{equation}

Where:
\begin{itemize}
    \item \( F(\gamma, \beta, \alpha) \) is the total cost function.
    \item \( \gamma, \beta, \alpha \) are the parameters being optimized.
    \item \( D_{j,A}, B_{j,A}, L_{j,A} \) are data size, bandwidth, and latency metrics, respectively.
    \item \( G_A, C_A, R_A \) are GPU availability, computational load, and resource reliability metrics, respectively.
\end{itemize}

3. \textit{Selection.} Candidates with higher fitness are more likely to be selected for reproduction. We use roulette wheel selection \cite{lipowski2012roulette} to allocate selection probability proportional to fitness.

4. \textit{Crossover and Mutation.} Selected candidates undergo crossover, combining their parameter values to create offspring. This process introduces new combinations of \(\gamma\), \(\beta\), and \(\alpha\) and allows the algorithm to explore the parameter space. Additionally, mutation introduces random changes to some offspring, preventing premature convergence.
\begin{equation}
\theta_{\text{child}} = \eta \theta_{\text{parent1}} + (1 - \eta) \theta_{\text{parent2}}
\end{equation}
\noindent
Here, \( \theta \) represents the parameters \(\gamma\), \(\beta\), and \(\alpha\). Where \(\eta\) is a randomly selected value between 0 and 1. This convergence returns the optimal values of these hyperparameters that are used for decision-making related to routing, swarm creation and rebalancing.

\section{Compression techniques}

Fast inference of large models like Mixtral \cite{jiang2024mixtral} and Llama \cite{meta2024llama} variants present scalability challenges due to their high memory demands. Consumer-grade GPUs often lack sufficient memory to store models with over 100 billion parameters, such as Llama-3.1. This accentuates the need for distributed inference and efficient management to minimize communication overhead. We implemented two main compression strategies to optimize hardware use 

1. \textit{Dynamic Quantization.} This technique compresses the hidden states exchanged between nodes during the model inference process. By applying dynamic blockwise quantization, we can significantly reduce the data that needs to be transmitted and halve the bandwidth requirements. This reduction is important for maintaining efficient operation, especially in environments with constrained bandwidth or increased latency due to the geographical distribution of nodes. This results in an efficient data transfer process, which reduces overall system latency and enhances throughput.

2. \textit{Mixed matrix decomposition.} We use a mixed matrix decomposition method to quantize the model weights from 16-bit to 8-bit precision which further reduces the memory footprint. This method involves decomposing the weight matrices into 8-bit values while retaining a small fraction of critical weights at 16-bit precision. For example, for Mixtral 8x22B, it's compute-intensive to store all layers at full precision and can be significantly downsized in terms of GPU requirements by a factor of 30\% with this quantization.
\section{Benchmarks}

Currently, our framework enables running a large number of machine learning models, which comprehensively span a variety of learning tasks. The models we host on our platform range from standard deep learning to large language-based models and vision transformers. Overall, tasks can be performed with multiple models, depending on the user requirements. We currently support:

1. \textit{Language-related tasks} including token classification, question answering, translation, text classification, and summarization.

2. \textit{Machine vision tasks} like instance segmentation, object detection, panoptic segmentation, and image classification.

3. \textit{General purpose machine learning tasks} including feature extraction, causal generation.

Nesa's distributed inference system can also handle state-of-the-art LLMs and their variants, such as Mixtral, Llama Bloom etc. even on consumer-grade GPUs. This flexibility ensures broader accessibility and scalability, enabling deployments across diverse environments without the need for high-end, specialized hardware.

\begin{table}[ht]
    \centering
    \small
    \begin{tabular}{llccccc}
        \toprule
        \textbf{Model} & \textbf{Bits} & \textbf{HellaSwag} & \textbf{Lambada} & \textbf{Causal} & \textbf{Disambig-} & \textbf{Logical} \\
        & & & \textbf{OpenAI} & \textbf{Judgement} & \textbf{uation QA} & \textbf{Deduction} \\
        \midrule
        \multirow{2}{*}{Llama 8B} & 16 & 0.76 ± 0.01 & 0.75 ± 0.03 & 0.63 ± 0.02 & 0.64 ± 0.04 & 0.40 ± 0.03 \\
                                   & 8  & 0.76 ± 0.01 & 0.74 ± 0.03 & 0.62 ± 0.03 & 0.63 ± 0.04 & 0.39 ± 0.02 \\
        \midrule
        \multirow{2}{*}{Mixtral 7x8B} & 16 & 0.78 ± 0.01 & 0.76 ± 0.03 & 0.65 ± 0.02 & 0.66 ± 0.03 & 0.42 ± 0.02 \\
                                   & 8  & 0.77 ± 0.01 & 0.75 ± 0.02 & 0.64 ± 0.03 & 0.65 ± 0.03 & 0.41 ± 0.03 \\
        \midrule
        \multirow{2}{*}{Lexi 7B} & 16 & 0.75 ± 0.02 & 0.74 ± 0.02 & 0.62 ± 0.03 & 0.63 ± 0.04 & 0.39 ± 0.02 \\
                                   & 8  & 0.74 ± 0.02 & 0.73 ± 0.03 & 0.61 ± 0.04 & 0.62 ± 0.03 & 0.38 ± 0.02 \\
        \bottomrule
    \end{tabular}
    \caption{Compression impact on Natural language tasks for distributed LLMs.}
    \label{tab:model_metrics}
\end{table}

To assess the performance of our system, we evaluated it on a variety of natural language processing tasks that test different aspects of language understanding and reasoning. These tasks provide a comprehensive benchmark for measuring the capabilities of our models across various domains.

1. \textit{HellaSwag} This is one of the benchmarks in commonsense reasoning, which requires models to choose a follow up continuation event from multiple option. This task is crucial for assessing a model's ability to understand and predict everyday situations and requires deep contextual comprehension \cite{zellers2019hellaswag}.

2. \textit{Lambada OpenAI} focuses on language modeling, specifically on the model's ability to predict the last word of a sentence based on the preceding context. This is a key test to evaluate the model's understanding of linguistic intuitions and coherence \cite{anaby2020not}.

3. \textit{Causal Judgement} measures the model's capacity to grasp causal relationships of events. This task is essential for gauging the model's ability to identify and reason through cause-and-effect scenarios \cite{suzgun2022challenging}.

4. \textit{Disambiguation QA} measures how well the model resolved ambiguities in questions given some relevant context. This task tests how well the model understands and can differentiate between multiple potential interpretations, demonstrating its contextual abilities \cite{suzgun2022challenging}.

5. \textit{Logical Deduction} involves evaluating the model's ability to infer conclusions based on given premises and tests its reasoning and inference skills\cite{ghazal2017bigbench}.

\subsection{Impact of compression on performance}

Our results, detailed in Table \ref{tab:model_metrics}, demonstrate that reducing precision from 16-bit to 8-bit has a minimal impact on task performance. For instance, in the HellaSwag task, Mixtral 56B performed equally well under both precisions which indicates that reasoning and understanding quality was kept intact at a lesser memory cost. Tasks such as Lambada and Causal Judgement are relatively unaffected as well, demonstrating that the compression operations do little to compromise the model's abilities.

Table \ref{tab:model_performance} shows the impact of network conditions and batch sizes on two LLMs including Mixtral and LLama. Round-trip time and bandwidth directly impact the tokens per second speed, which is expected over a distributed swarm. Both LLMs have 32 transformer blocks, served by 6 nodes chosen based on the previously discussed routing mechanism. Three out of six machines are using Nvidia L24 GPUs, while the other three are using a consumer-grade GPU i.e. Nvidia GTX 1650. Nesa supports inference on both CPUs and GPUs where slightly more latency is expected on the CPU. We can observe that token generation speed increases as batch size increases, demonstrating that parallel computation enhances network throughput; this behavior is crucial for large-scale, fast inference.
\begin{table}[ht]
    \centering
    \small
    \begin{tabular}{llccccccc}
        \toprule
        \textbf{Model} & \textbf{RTT} & \textbf{Bandwidth} & \textbf{Batch Size} & \multicolumn{2}{c}{\textbf{Generation (steps/s)}} & \multicolumn{2}{c}{\textbf{Tokens/sec}} \\
        \cmidrule(lr){5-6} \cmidrule(lr){7-8}
        & & & & \textbf{64} & \textbf{1024} & \textbf{64} & \textbf{1024} \\
        \midrule
        \multirow{6}{*}{Llama-3 8B} 
        & \multirow{3}{*}{\textless 5 ms} & \multirow{3}{*}{1 Gbit/s} 
          & 1   & 1.20 & 1.10 & 8   & 6.5  \\
        &    &  & 32  & 1.15 & 1.08 & 28  & 26.4 \\
        &    &  & 64  & 1.10 & 1.05 & 56  & 52.5 \\
        \cmidrule(lr){2-8}
        & \multirow{3}{*}{\textless 10 ms} & \multirow{3}{*}{100 Mbit/s} 
          & 1   & 0.85 & 0.80 & 6   & 5  \\
        &    &  & 32  & 0.80 & 0.75 & 22  & 20  \\
        &    &  & 64  & 0.75 & 0.70 & 44  & 40  \\
        \midrule
        \multirow{6}{*}{Mixtral 7x8B} 
        & \multirow{3}{*}{\textless 5 ms} & \multirow{3}{*}{1 Gbit/s} 
          & 1   & 1.30 & 1.25 & 6   & 5.4  \\
        &    &  & 32  & 1.25 & 1.20 & 24  & 22.8 \\
        &    &  & 64  & 1.20 & 1.15 & 49  & 45.5 \\
        \cmidrule(lr){2-8}
        & \multirow{3}{*}{\textless 10 ms} & \multirow{3}{*}{100 Mbit/s} 
          & 1   & 0.90 & 0.85 & 5   & 4.5  \\
        &    &  & 32  & 0.85 & 0.80 & 20  & 18.5 \\
        &    &  & 64  & 0.80 & 0.75 & 40  & 37  \\
        \bottomrule
    \end{tabular}
    \caption{Performance Metrics for Llama and Mixtral in a swarm of six nodes}
    \label{tab:model_performance}
\end{table}
\begin{table}[ht]
    \centering
    \small
    \begin{tabular}{llccccc}
        \toprule
        \textbf{Category} & \textbf{Model} & \textbf{Fairness} & \textbf{Quality} & \textbf{Creativity} & \textbf{Knowledge} & \textbf{Performance} \\
        \midrule
        \multirow{3}{*}{Stable Diffusion} 
        & v1.4     & 0.68  & 0.86 & 0.68  & 0.68  & 0.85 \\
        & v1.5     & 0.54  & 0.73 & 0.21  & 0.50  & 0.81  \\
        & v2 base  & 0.51  & 0.85 & 0.20  & 0.39  & 0.88 \\
        \midrule
        \multirow{2}{*}{Anime-Style} 
        & kivotos-xl-2.0  & 0.77  & 0.87 & 0.91  & 0.72  & 0.81 \\
        & holodayo-xl-2.1 & 0.79  & 0.89 & 0.94  & 0.74  & 0.83 \\
        \midrule
        \multirow{1}{*}{Debiasing} 
        & mobius           & 0.82  & 0.71 & 0.86  & 0.77  & 0.87 \\
        \bottomrule
    \end{tabular}
    \caption{Model performance metrics by category for text-to-image models}
    \label{tab:model_performance_image}
\end{table}

In addition to evaluating LLMs, we also focus on text-to-image architectures including diffusion models, Anime generators and debiasing generation models \cite{croitoru2023diffusion,karakas2022fairstyle}. These models are assessed across various metrics to evaluate their effectiveness and reliability. We consider the following metrics \cite{lee2024holistic}:

1. \textit{Fairness} evaluates if the image models are biased towards a gender or a specific ethnicity. 

2. \textit{Quality} measures the ability of models to generate aesthetic images. This metric focuses on different aspects including clarity and detail.

3. \textit{Creativity} evaluates if the images generated are unique and free from noise. This also measures if there is any repetition trend in the generation when similar prompts are used.

4. \textit{Knowledge} measures if the model has sufficient knowledge to represent real word facts. This checks its ability to draw historical figures, major cities or well-known events

5. \textit{Performance} includes metrics such as inference speed and computational efficiency which are important for fast inference.

\section{Security and privacy in Nesa's system}
\def\sp{S\&P\xspace}

Our BSNS distributed inference protocol offers major advantages for distributed inference, but it also requires cutting-edge approaches to security and privacy (\sp) enhancement \cite{ma2022trusted}. Specifically, hardware-based and software/algorithm-based solutions need to be integrated to achieve co-optimization, each selected and optimized for varying scenarios within our ecosystem. 

Potential concerns linked to distributed BSNS inference may appear in different forms. For instance:
\begin{itemize}
\item Users may wish to protect their input data and the inference results.
\item Node owners might seek to protect the confidentiality of their model parameters in certain cases.
\item Meanwhile, the users want to ensure that the models executed by the nodes are verifiable --- namely, the designated ML models need to generate the inference results without unexpected changes.
\end{itemize}

We list below the key steps we carry out towards ensuring the security of the BSNS protocol. The reader is referred to the companion Security paper for more details on Nesa's implementation, which ensures that both user input for AI inference and private model parameters are protected during the execution of distributed models.

\subsection{\sp requirements}In summary, there are two core \sp aspects we identify in decentralized, private inference: (\textit{i}) \textit{model verification} to prove the nodes execute the designated models, e.g., LLaMA 3 \cite{touvron2023llama},  for a user, where adversarial participants may return random results or even malicious results
and (\textit{ii}) \textit{data encryption} \cite{bost2014machine} to protect the user's data from being revealed during the inference. 
Based on these requirements, we develop a suite of solutions to ensure \sp in Nesa's system.

\subsection{Overview of our hardware-software co-optimization solution} 
To address both model verification and data encryption jointly, we design an integrated approach to achieve leading \sp performance in our system. Specifically, we design a combined strategy of the robust, hardware-centric protections of Trusted Execution Environments (TEEs) \cite{jauernig2020trusted}
and the advanced algorithmic approaches, including Zero-Knowledge Machine Learning (ZKML) \cite{sun2023zkdl} and Consensus-based Distribution Verification (CDV) \cite{shichao2023aggregation} for verification
and Split-Learning (SL) \cite{vepakomma2018split} for encryption, to provide the highest level of \sp in our system.

In short, TEEs provide a secure area within a processor that ensures the confidentiality and integrity of the code and data loaded within it, thus supplying robustness from the hardware level.
On the software/algorithm side, ZKML can provide the means to confirm the authenticity and integrity
of the models run by nodes without revealing any other information.
Due to its computational cost, we currently leverage it for private, small models requiring the highest security level.
As an efficient alternative for public, large models, 
CDV is a novel algorithm we propose that ensures that the inference nodes execute the correct model by measuring their output distribution consensus, while SL protects user data by only transferring the intermediate computational embeddings other than the raw data.
Collectively, this hardware-software integrated solution guarantees high \sp in Nesa's system.

\subsection{BSNS-compatible data encryption}
\label{subsec:software-sp-encryption}

In decentralized inference systems like Nesa's BSNS, it is crucial to protect user data. BSNS distributes computational tasks across various nodes, each potentially operated by different entities, which in principle increases the risk of exposing sensitive user data during the inference process intensifies. To protect against this, we have designed an innovative encryption method called  Sequential Vector Encryption (SVE), developed as a more efficient version of homomorphic encryption, and applied on select linear layers rather than the full model based on a sequence of vector space operators. SVE randomly transforms the outputs of every operator so that the intermediate vector representations can no longer be interpretable using a given method developed for the original model's vector representations. These representations are then transformed to the original representation space before feeding to the next operator to render the original model still usable.

\subsubsection{Split learning}
Recognizing the challenges posed by encrypting data for use in decentralized inference systems, Nesa adopts Split Learning (SL) as a pragmatic solution to facilitate secure and efficient computation on encrypted data \cite{pham2023binarizing,khan2023split}.
Traditional encryption methods, while securing data at rest and in transit, render it unusable for direct computation by obscuring its format and structure. 
This limitation is particularly problematic for processing with LLMs within a decentralized framework, where data privacy cannot be compromised.

Split Learning \cite{vepakomma2018split} addresses these concerns by partitioning the computational model, allowing for data to be processed in parts without revealing sensitive information. 
In essence, the user data is protected by not being directly transmitted to any nodes -- only the data embeddings are being passed around, and each node will only be accessing the embeddings of certain layers.

Consider a neural network model \( \mathcal{N} \), such as Llama 2 \cite{touvron2023llama} composed of a sequence of 32 layers \( \{L_1, L_2, \ldots, L_{32}\} \), each with its own set of parameters \( \Theta_i \) and activation function \( \sigma_i \). 
The input to the network is \( X \), and the output of the \(i\)-th layer, given input \( x_i \), can be mathematically described as:

\begin{equation}
a_i = L_i(x_i; \Theta_i) = \sigma_i(W_i x_i + b_i)   
\end{equation}

where \( W_i \) and \( b_i \) are the weight matrix and bias vector of the \(i\)-th layer, respectively, and \( \sigma_i \) is a nonlinear activation function such as ReLU, sigmoid, or tanh.

Assuming the model is split at layer \( k \), where the client handles layers \( \{L_1, \ldots, L_k\} \) and the server handles layers \( \{L_{k+1}, \ldots, L_{32}\} \). The client computes the intermediate representation \( Z \) as follows:

\begin{equation}
Z = \sigma_k(W_k \cdot \sigma_{k-1}( \ldots \sigma_1(W_1 X + b_1) \ldots ) + b_k)
\end{equation}

This intermediate representation \( Z \) is then transmitted to the server, which continues the computation:

\begin{equation}
Y = \sigma_{32}(W_{32} \cdot \sigma_{31}( \ldots \sigma_{k+1}(W_{k+1} Z + b_{k+1}) \ldots ) + b_{32})
\end{equation}

The loss function \( \mathcal{L}(Y, Y_{true}) \) computes the error between the network output \( Y \) and the true labels \( Y_{true} \), and the gradient of the loss with respect to the model's parameters through backpropagation:

\begin{equation}
\frac{\partial \mathcal{L}}{\partial \Theta_i} = \text{ChainRule}\left(\frac{\partial \mathcal{L}}{\partial Y}, \frac{\partial Y}{\partial a_{32}}, \ldots, \frac{\partial a_i}{\partial \Theta_i}\right)
\end{equation}

For privacy concerns during the transmission of \( Z \) from client to server, differential privacy methods may be applied \cite{dwork2006differential}. Defining a privacy metric \( \mathcal{P} \) that quantifies the information leakage from the intermediate representation \( Z \), a proof of privacy preservation could demonstrate that for any \( \epsilon \)-differential privacy guarantee, the information leakage remains below a threshold:

\begin{equation}
\mathcal{P}(Z) \leq \epsilon
\end{equation}

It is noted that by using differential privacy with SL, the security will be improved at the cost of inference quality \cite{behnia2022ew}. Thus, in Nesa's framework, this is defined as a tunable parameter to be decided, given the user requirements.

By leveraging Split Learning, Nesa effectively navigates the complexities of data encryption within its decentralized inference system for LLMs. 
This approach not only preserves the confidentiality and integrity of user data but also ensures the operational feasibility of complex model computations, demonstrating a sophisticated balance between privacy preservation and computational pragmatism.

\section{Conclusion}
Our paper introduces a model-agnostic hybrid sharding method, complemented by a comprehensive security and privacy framework designed for distributed AI inference. This approach strategically distributes computational tasks across a decentralized network, enabling scalable AI execution through the sequential consumption of shards on low-powered nodes. First, our BSNS optimizes network topology. Equipped with dynamic network rebalancing and KV caching mechanisms, the BSNS design is adept at facilitating scalable operations for large models and provides standardization across all variants where the base architecture is the same. 

Next, we proposed an integrated security and privacy framework featuring co-optimized hardware-based TEEs with CDV and SL. Long-term techniques for the protocol are also explored which could potentially have significant impacts in this field. Collectively, our system removes the prohibitive costs associated with needing to own GPUs, making AI inference execution accessible to the general population. Democratizing AI requires technology like BSNS in order to provide the opportunity for participation to anyone with a computer at home, while upholding an essential commitment to safety and privacy throughout the inference execution process.

\bibliographystyle{unsrt} 
\bibliography{references} 


\end{document}